\documentclass[letterpaper, 10 pt, conference]{ieeeconf}  

\IEEEoverridecommandlockouts                              

\usepackage{cite}
\usepackage{amsmath,amssymb,amsfonts}
\usepackage[ruled,vlined,linesnumbered,noend]{algorithm2e}
\usepackage{graphicx}
\usepackage{textcomp}
\usepackage{xcolor}
\usepackage{makecell}
\usepackage{amsmath}
\usepackage{float}
\usepackage{multirow}
\usepackage[font=footnotesize]{caption}
\usepackage{subcaption}

\usepackage{url}  

\begin{document}
\title{\Large \bf
In-Hand Manipulation of Unknown Objects with Tactile Sensing for Insertion
}

\author{Chaoyi Pan$^{*}$, Marion Lepert$^{*}$, Shenli Yuan, Rika Antonova, Jeannette Bohg 
\thanks{
$^{*}$Authors contributed equally and names are in random order. Chaoyi Pan is with Tsinghua University. Marion Lepert, Shenli Yuan, Rika Antonova, and Jeannette Bohg are with Stanford University {\tt\small [lepertm, shenliy, rika.antonova, bohg]@stanford.edu}. \ Toyota Research Institute provided funds to support this work. We are grateful to Shaoxiong Wang for help with the Roller Grasper. Rika Antonova is supported by the National Science Foundation grant No.2030859 to the Computing Research Association for the CIFellows Project.
}}

\pdfinfo{
   /Author (Chaoyi Pan, Marion Lepert, Shenli Yuan, Rika Antonova, Jeannette Bohg)
   /Title  (In-Hand Manipulation of Unknown Objects with Tactile Sensing for Insertion)
   /CreationDate (D:2022101020000)
   /Subject (Robotics)
   /Keywords (In-hand manipulation, Bayesian optimization, Tactile sensing)
}



\maketitle
\thispagestyle{empty}
\pagestyle{empty}

\begin{abstract}

In this paper, we present a method to manipulate unknown objects in-hand using tactile sensing without relying on a known object model.
In many cases, vision-only approaches may not be feasible; for example, due to occlusion in cluttered spaces. We address this limitation by introducing a method to reorient unknown objects using tactile sensing. It incrementally builds a probabilistic estimate of the object shape and pose during task-driven manipulation. Our approach uses Bayesian optimization to balance exploration of the global object shape with efficient task completion. To demonstrate the effectiveness of our method, we apply it to a simulated Tactile-Enabled Roller Grasper, a gripper that rolls objects in hand while collecting tactile data. We evaluate our method on an insertion task with randomly generated objects and find that it reliably reorients objects while significantly reducing the exploration time. 

\end{abstract}

\section{INTRODUCTION}
This work studies how robots can reorient objects in-hand with limited prior knowledge about object shape. Manipulating objects of unknown shapes is a foundational skill 
to perform tasks in unstructured environments. For example, to be useful in a kitchen, a robot must be able to pick up and manipulate a breadth of objects such as fruits and vegetables whose shapes cannot be known beforehand and fit them in constrained spaces such as a tightly packed drawer in a fridge. While vision is a powerful modality to gain information about novel objects, there are many scenarios where vision cannot be used, such as cluttered spaces with high occlusion and during manipulation of small objects where the gripper will often occlude the object. Therefore, robots must learn to manipulate objects with sensors other than vision. Tactile sensing provides high-resolution local information about contact between the object and gripper, which is complementary to global vision information. 

Object reorientation with tactile sensing is challenging because tactile data gives information about the object shape for only a small contact patch area, so the object’s global shape needs to be pieced together from limited local information. In addition, the object may only have a limited set of features that the tactile sensor can detect to distinguish between different locations on the object. Moreover, the object may shift slightly in-between tactile readings, which increases uncertainty regarding the location of these readings.  
\begin{figure}[t]
    \vspace{5px}
    \centering
    \includegraphics[width=\linewidth]{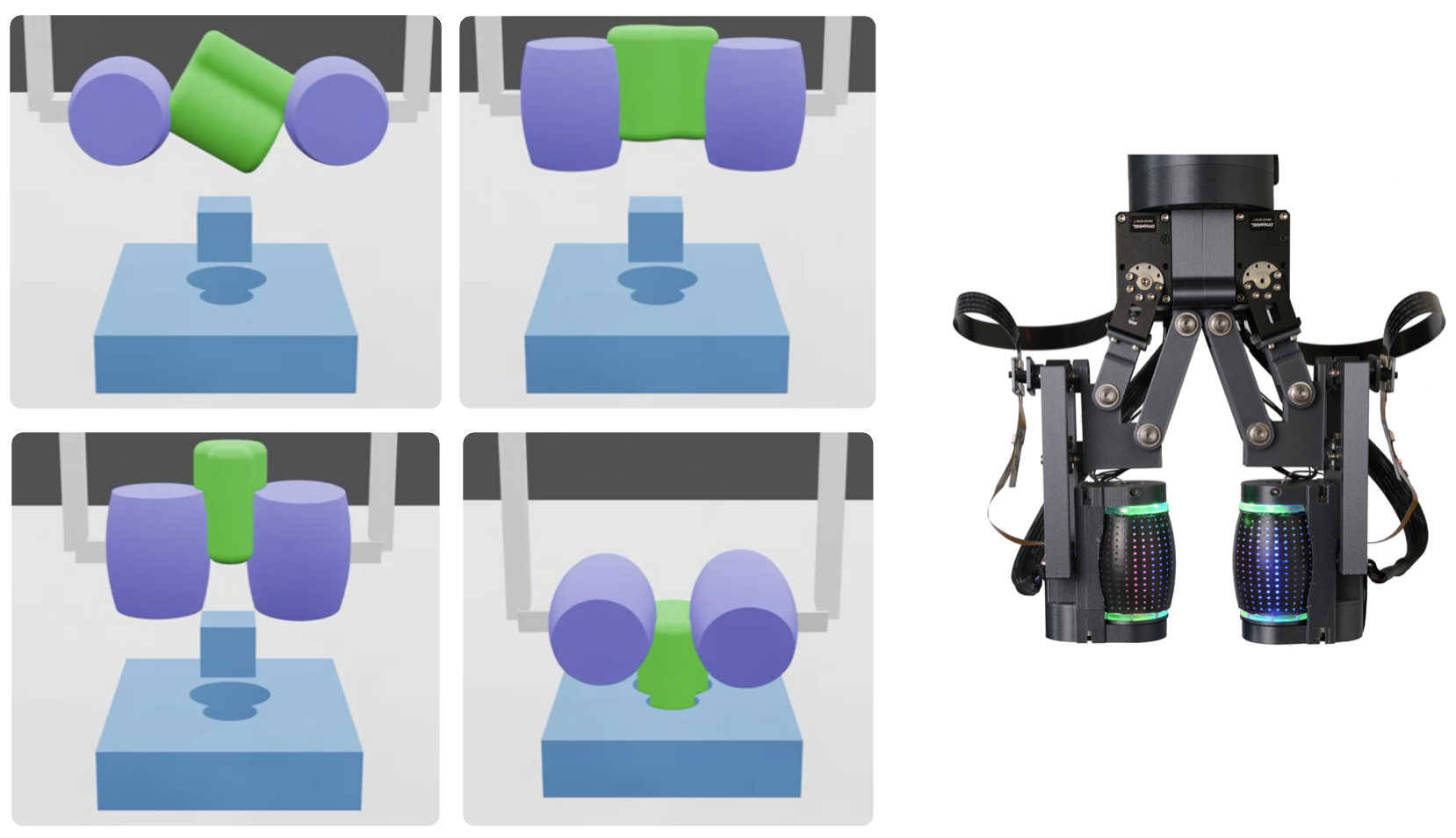}
    \vspace{-15px}
    \caption{Left: Simulation of the Tactile-Enabled Roller Grasper demonstrating a sequence of exploration steps that lead to successful insertion. Right: The Tactile-Enabled Roller Grasper that we simulate and that inspired our work.}
    \label{fig:roller_grasper_hardware}
    \vspace{-20px}
\end{figure}
We seek to overcome some of these challenges by leveraging the Tactile-Enabled Roller Grasper. This gripper rolls objects in hand and continuously collects tactile data on the surface of the rollers. Staying in contact with the object reduces uncertainty between tactile measurements, and enables us to piece together a sequence of local contact patches into a global estimate of the object's shape. 
Given this gripper, we propose a method to reorient unknown objects by incrementally building a probabilistic estimate of the object's shape during manipulation. Our method leverages Bayesian optimization to strategically trade off exploration of the global object shape with efficient task completion. We demonstrate our approach on a simulated Tactile-Enabled Roller Grasper as shown in Fig. \ref{fig:roller_grasper_hardware}. 

We evaluate our approach on an insertion task. Insertion tasks are ubiquitous in many environments, such as assembly tasks in factories, dense box packing in warehouses, and plugging cables in the home. As a result, insertion tasks continue to be heavily studied in robotics. In this paper, we focus on finding the correct object orientation that will allow the object to fit into a target hole. The robot has access to a parameterization of the hole's contour, which gives the robot a well defined reorientation target without revealing the 3D-shape of the object, ensuring that our assumption that the robot is working with an unseen object holds. We evaluate our method in simulation on a set of randomly generated objects and find that our method reliably completes the insertion task while significantly reducing the exploration time needed to do so. 

To summarize, our main contributions are: 
(i) A system to reorient unknown objects that does not rely on vision and instead leverages tactile sensing; and (ii) An in-hand 3D simultaneous shape reconstruction and localization method to estimate an object’s shape and pose that is task driven.

\section{RELATED WORK}
\subsection{In-hand manipulation}
\label{RelatedWorkInhandManipulation}


In-hand manipulation has been studied extensively \cite{andrychowicz2020learning, chavan2018hand, yuan2020design, 8950086, fingergaiting, bhatt2022surprisingly, chen2022system}. Many approaches require object pose which can be obtained with a marker tag \cite{yuan2020design,8950086} or a 6D object pose estimator \cite{fingergaiting}. Alternatively, deep learning methods can obtain an end-to-end policy that does not require  pose estimation. For example, \cite{andrychowicz2020learning} learns a controller to reorient a Rubik's cube and \cite{chen2022system} presents a controller that can reorient many distinct objects, but both rely on vision and complex multi-fingered grippers. In contrast, \cite{chavan2018hand} demonstrates how simpler hardware (parallel-jaw grippers) can use extrinsic dexterity to re-orient objects in-hand. Their approach is limited to simple cuboids and requires a 6D object pose estimator. \cite{bhatt2022surprisingly} shows how a compliant gripper can robustly reorient objects in-hand using handcrafted open-loop primitives that do not require object pose or shape estimation.

\subsection{Tactile Sensing for shape reconstruction and localization}
\label{RelatedWorkTactileSensing}

Reconstructing object shape from vision and tactile data is a common research objective \cite{Falco, suresh2022shapemap, smith20203d}. These works typically rely heavily on vision and tend to use tactile data simply to detect contact. However, many scenarios with heavy occlusion  preclude good vision data, leading to a growing interest in reconstructing and/or localizing objects with just tactile data. 

Low resolution tactile sensors \cite{schaeffer2003methods, petrovskaya2011global, driess2017active} are typically used to obtain binary contact information only. The development of vision-based high resolution tactile sensors, such as GelSight \cite{li2014localization}, has greatly increased our ability to reconstruct and localize objects without vision. The GelSight outputs an RGB image of the tactile imprint, and photometric stereo is used to convert this image to a depth image. Others learn this mapping from data \cite{bauza2019tactile}, or choose instead to learn a binary segmentation mask \cite{bauza2020tactile} to reduce noise.  \cite{li2014localization} showed one of the first uses of the GelSight  for small object localization, further improved in \cite{Luo, bauza2019tactile}. However, they require building a complete tactile map of the object before using it for localization.  

Most works assume each tactile imprint is collected by making and breaking contact with the object. This introduces uncertainty in the relative position of tactile imprints. Instead, \cite{driess2017active} maintains constant contact with the object to reduce robot movement and obtain higher fidelity data. However, they use a low resolution sensor and only consider a bowl shaped object fixed in space. \cite{she2021cable} slide along a freely moving cable with a GelSight to estimate the cable's pose, but their method is specific to cables. By rolling objects in-hand, we also continuously collect tactile data, but do so for a more general set of freely-moving objects.

Most of these works narrow their scope either by reconstructing the shape of an unknown object in a fixed pose, or by localizing an object of known shape and unknown pose. Recently, \cite{suresh2021tactile} proposed removing these limitations by simultaneously doing shape reconstruction and localization from tactile data. However, they only show results for pushing planar objects. 


Gaussian process implicit surfaces (GPIS) are commonly used to build a probabilistic estimate of an object's shape \cite{5980395}. \cite{bjorkman2013enhancing, driess2017active} use the variance of the GPIS to guide the exploration towards regions with highest uncertainty. However, the goal of these approaches is to reconstruct the object shape as accurately as possible. In contrast, we focus on exploration that prioritizes regions of the object that are likely to aid in solving an insertion task, making our exploration more efficient. \cite{dragiev2013uncertainty}  also prioritizes exploring regions that help solve a task, but their approach is limited to grasping an object that is fixed during exploration.



\subsection{Insertion Task}

A lot of previous work on the peg insertion task is object-geometry specific \cite{bruyninckx1995peg, chhatpar2001search, park2017compliance}. For example, \cite{peg1} assumes a cylindrical peg shape;  \cite{michelle48} uses vision and tactile data to enable insertion of complex shapes but assumes known peg geometry; \cite{peg2} learns from forces measured during human demonstration, but requires successful demonstration of the specific peg shape. In contrast, our work assumes no prior knowledge of the peg shape, and instead learns an explicit peg object model. There is some prior work that aims to be robust to different peg geometries, such as \cite{Michelle}, but the pegs are fixed to the end-effector. 

\subsection{Roller Grasper}
In this work, we use a simulated version of the Tactile-Enabled Roller Grasper hardware from \cite{yuanthesis}. This gripper has 7 degrees of freedom, shown in Fig. \ref{fig:GO}, and each roller is equipped with a custom GelSight sensor \cite{yuan2017gelsight}. The sensor's camera is inside the roller, fixed to the stator such that it points towards the grasping point, while the elastomer covers the rotating roller.

Because the Roller Grasper rolls objects in-hand, it can continuously collect tactile information of an object's surface. This continuity reduces uncertainty in relative position between tactile imprints, simplifying reconstruction of object shape compared to linkage-based grippers that capture discontinuous data.

Several works have studied the Roller Grasper. \cite{yuan2020design} proposed a velocity controller and an imitation learning controller to reorient objects. \cite{yuanthesis} built a closed loop controller that keeps an object centered in the Roller Grasper during manipulation using contact patch data from the tactile sensor. We build on these works by using the velocity controller to determine the roller action to produce a desired object angular velocity. However, we do not assume a known object model or use marker tags, and instead simultaneously reconstruct and localize the object in a task-driven manner.

\vspace{10px}
\section{APPROACH}
\vspace{3px}
We propose a method that leverages tactile sensing to reorient objects of unknown shapes in order to complete a task. We focus on settings with high occlusion where the robot cannot see the object it is manipulating, and must instead rely on its sense of touch and proprioception. We evaluate our method on an insertion task where the robot must reorient an object with a set of small and easily occluded features to fit into a hole. Our approach consists of three parts: shape estimation, task-guided exploration and in-hand reorientation. The shape estimation and task guided exploration are performed iteratively before the final in-hand reorientation.
\begin{figure}[t]
    \vspace{7px}
    \centering
    \includegraphics[width=0.98\linewidth]{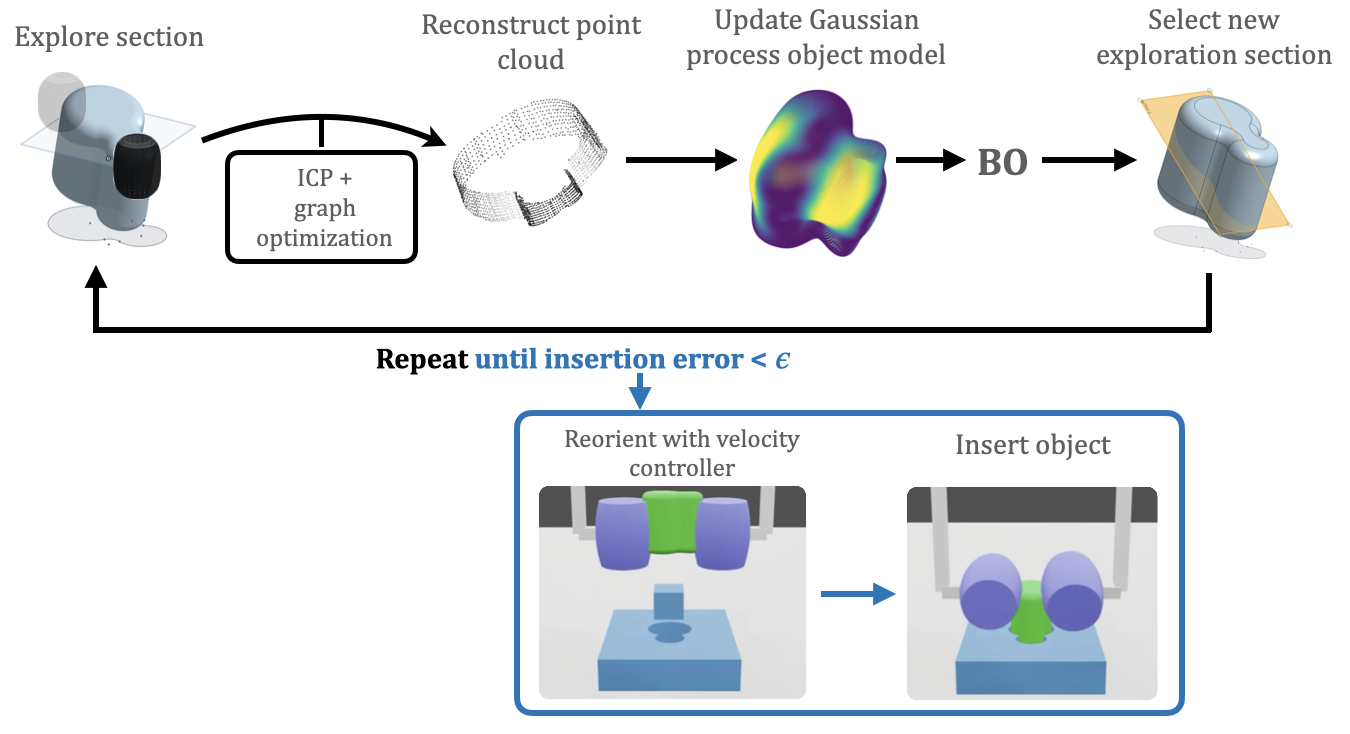}
    \caption{A high-level overview of our method.}
    \label{fig:hardware_icp}
    \vspace{-15px}
\end{figure}
\begin{itemize}
    \item \textbf{Shape estimation}: The robot collects proprioception and tactile data as it rolls the object in-hand. We use the Iterative Closest Point (ICP) method \cite{rusinkiewicz2001efficient} and graph optimization \cite{choi2015robust} to locally align the tactile data and estimate the global shape of the object along the touched contours. We fit a Gaussian process to this data to estimate the whole shape of the object. 
    \item \textbf{Task guided exploration}: We decompose the insertion task into a form that allows us to leverage Bayesian optimization~\cite{jones1998efficient, BOtutorial2016}. Thereby, we can guide object shape exploration to be in service of the insertion task and avoid unnecessary exploration that is unlikely to aid in task completion.
    \item \textbf{In-hand reorientation}: Once our exploration algorithm has high-confidence that it has found an orientation of the object that will fit in the hole, we localize the estimated object model using tactile data and use a velocity controller to reorient the object into the hole.
\end{itemize}

\textbf{Assumptions}: Because the Roller Grasper is physically incapable of reorienting objects with large aspect ratios without re-grasping, and because re-grasping is beyond the scope of this paper, we assume that the object to be manipulated has an aspect ratio close to 1. To enable the Roller Grasper to accurately reconstruct and localize objects, we also assume that our object has small features that can be detected by the tactile sensor which enable finding corresponding points on the object surface. We assume that an approximate initial placement of the object is known to allow us to easily grasp the object, since we focus on in-hand manipulation rather than details of grasping. Since the Roller Grasper is position controlled, we assume we know the approximate width of the object to set the target distance between the rollers. This could be overcome by using a force sensor at each roller that would allow the grasper to regulate a desired grasping force.

\subsection{Local shape estimation from tactile images}
\label{SectionShapeEstimation}

To estimate object shape, the Roller Grasper rolls the object in-hand and gathers joint position, $X \in \mathbb{R}^7$, and a tactile height map, $I \in \mathbb{R}^{W\times H}$, at each time step.

We approximate the shape of the object by using the joint position data to estimate the transformation between the depth maps. Because the object occasionally slips on the roller surface, these estimates are noisy. We use ICP between local pairs of depth maps to reduce this noise and improve the fidelity of the reconstruction.

\begin{figure}[t]
    \vspace{5px}
    \centering
    \includegraphics[width=8cm]{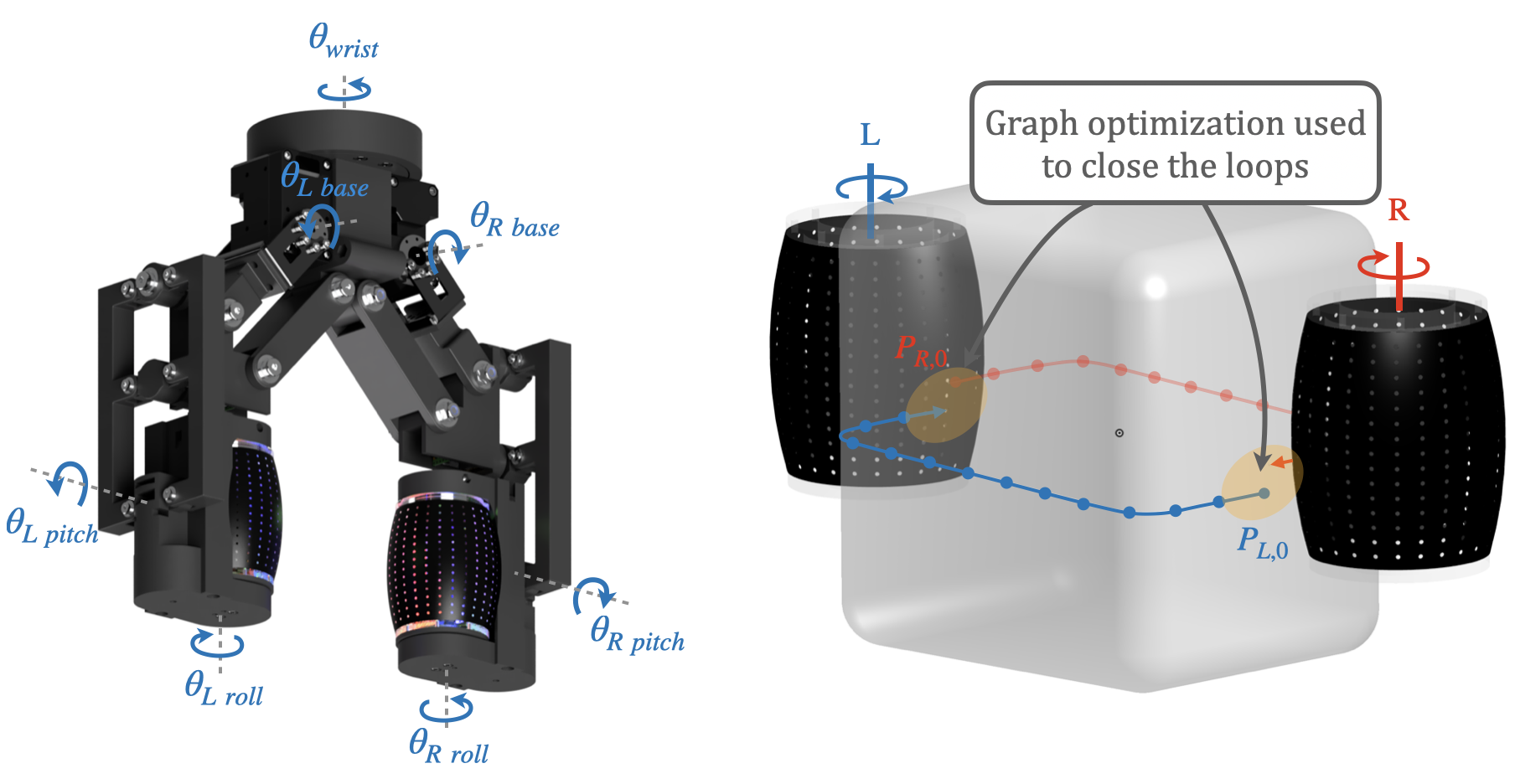}
    \caption{Left: Degrees of freedom of the Roller Grasper. Right: After the object rotates 180º, we use graph optimization to close the loop between the data collected by the left roller (blue) and the right roller (red).}
    \label{fig:GO}
\end{figure}

Additionally, once the object has rotated at least 180 degrees, the rollers encounter regions of the object that have already been scanned by the other roller, as shown in Fig.~\ref{fig:GO}. When we detect this loop closure, we use graph optimization \cite{choi2015robust} to align the two sets of point clouds. 

\subsection{Global shape representation with Gaussian processes} \label{GaussianShapeEstimate}
Because the proprioception and tactile data collected by the Roller Grasper only give us partial information about the object shape, we use a Gaussian process (GP) to build a probabilistic estimate of the overall shape.
We parameterize the surface of the object by spherical coordinates $\theta, \phi, r$. We use a GP to model the function $f(\theta, \phi) \!=\! r$ that represents the distance from the center of the object to its surface along a ray parameterized by $\theta$, $\phi$. This GP model lets us predict the mean $\bar{f}(\pmb{x}_*)$ and variance $\mathbb{V}[f(\pmb{x}_*)]$ of the distance to the object's surface along any `query' ray $\pmb{x}_* \!:=\! (\theta_*, \phi_*)$.

Formally, $\mathcal{GP}\big(m(\cdot), k(\cdot,\cdot)\big)$ is a model defined by a prior mean function $m(\cdot)$ and kernel function $k(\cdot, \cdot)$. The prior $m(\cdot)$ is usually taken as zero. The kernel encodes similarity between inputs: a large $k(\pmb{x}_i, \pmb{x}_j)$ implies that observing the value $r_i = f(\pmb{x}_i)$ for input $\pmb{x}_i$ would have a large influence on our estimate for $f(\pmb{x}_j)$ for input $\pmb{x}_j$.
We can compute the posterior mean and variance using:
\begin{align}
    \bar{f}_* &:= \bar{f}(\pmb{x}_*) := \pmb{k}_*^T (K + \sigma^2_nI)^{-1}\pmb{r}, \\ 
    \mathbb{V}[f_*] &:= \mathbb{V}[f(\pmb{x}_*)] := k(x_*,x_*) - \pmb{k}_*^T(K+\sigma_n^2I)^{-1} \pmb{k}_*
\end{align}
We obtain $\pmb{r}, K, \pmb{k_*}$ from the $N$ points of the object's point cloud $\{\pmb{x}_i \!:=\! (\theta_i, \phi_i), r_i\}_{i=1..N}$ reconstructed so far. $\pmb{r} \in \mathbb{R}^N$ is a vector with distances from the object's center to its surface (with entries $r_{i, i = 1..N}$). $K \!\in\! \mathbb{R}^{N \!\times\! N}$ is a matrix with entries $k(\pmb{x}_i, \pmb{x}_j)_{i,j=1..N}$; $\pmb{k}_* \!\in\! \mathbb{R}^{N}$ is a vector with entries $k(\pmb{x}_*, \pmb{x}_i)$.

\begin{figure}[t]
    \centering
    \vspace{7px}
    \includegraphics[width=8.7cm]{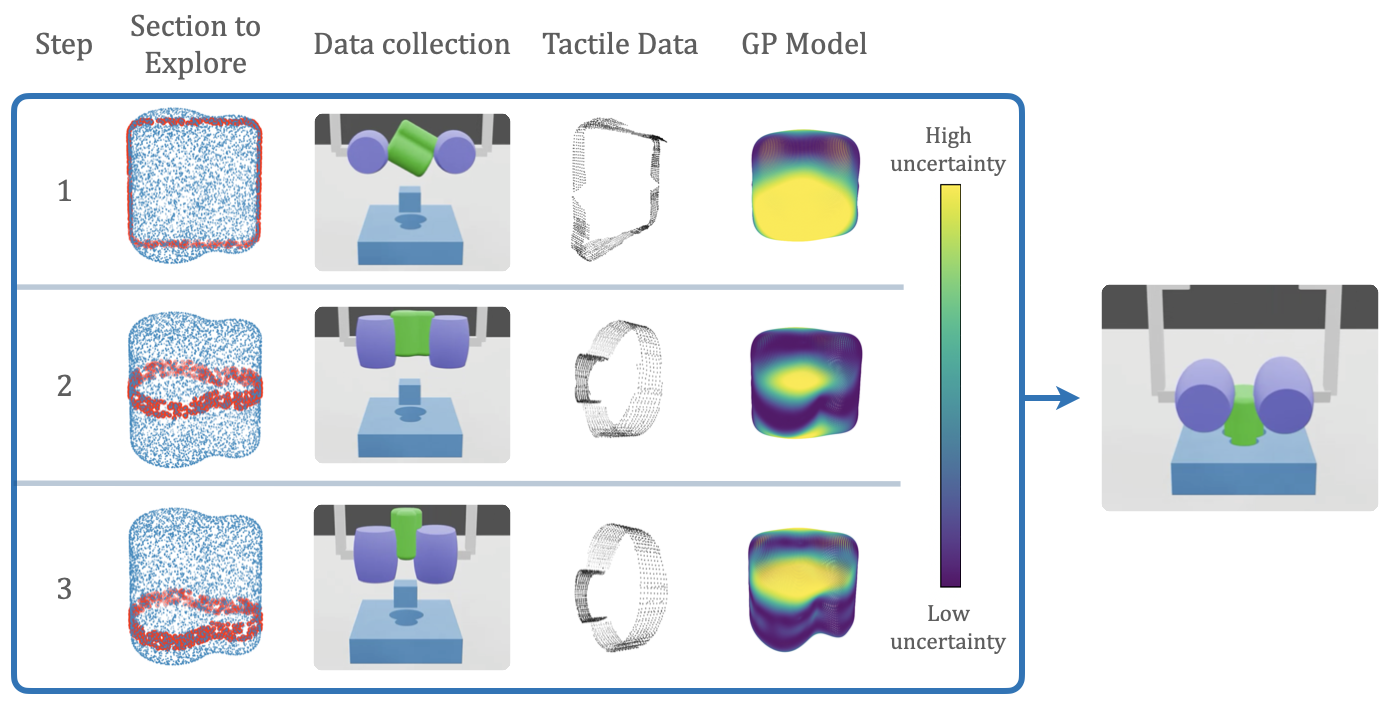}
    \caption{A demonstration of the reorientation procedure, where each row corresponds to one exploration step. The first column shows the section the roller selects to explore. The second column shows the simulated Roller Grasper rolling the object in hand along the selected section. The third column demonstrates the collected point cloud from the tactile images after using ICP and graph optimization. The fourth column shows the probabilistic object model described by the GP and trained from the collected tactile data.}
    \label{fig:appr:demo}
    \vspace{-10px}
\end{figure}

We use the Squared Exponential kernel function and leverage the GP marginal likelihood
to estimate the kernel hyperparameters and the noise parameter $\sigma_n$ automatically. See~\cite{gpml} for further details. 
Fig. \ref{fig:appr:demo} shows an example of incrementally building an object's shape estimate with a GP.

\subsection{Task Oriented Exploration}
We use {\em Bayesian Optimization\/} (BO) to guide our exploration of the object shape in order to efficiently identify  regions of the object that are important for completing the insertion task. We construct a task oriented \textit{acquisition} function for BO that selects the next target cross section of the object to explore during in-hand manipulation. The aim is to select targets to reduce the uncertainty over the object shape globally, but focus on the promising regions and avoid over-exploring parts of the object unlikely to be useful for successful insertion. In contrast to related work that focuses on uniformly minimizing uncertainty for overall shape reconstruction, we use the task objective to achieve targeted exploration. Below we describe how we construct the task oriented acquisition function for BO.

\begin{figure}[b]
    \centering
    \includegraphics[width=5cm]{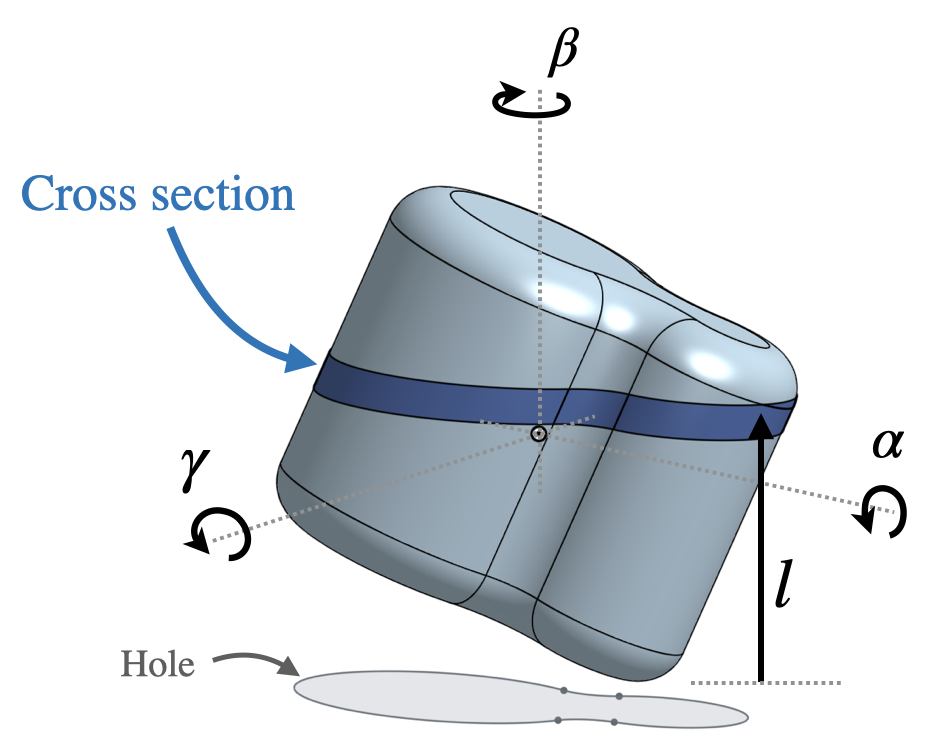}
    \caption{Each object cross section is parameterized by angles $\alpha$, $\beta$, $\gamma$ describing the orientation of the object relative to the hole and the height $l$. All cross sections have uniform thickness.}
    \label{fig:CrossSectionParameterization}
\end{figure}

We represent the orientation of the object with a discrete set of euler angles $\alpha$, $\beta$, $\gamma$. Our goal is to evaluate the likelihood of insertion for each orientation. To do this, we start with a probabilistic estimate of the object model. This model is a {\em Gaussian Process\/} (GP) described in Section~\ref{GaussianShapeEstimate}, which allows us to obtain estimates along any ray $\theta, \phi$ of the distance to the object surface $R(\theta, \phi) \sim \mathcal{N}(\bar{f}, \mathbb{V}[f])$ in a spherical coordinate system. For each orientation, we decompose the GP object model into horizontal cross sections of uniform width parameterized by height $l$ as 
shown in Fig.~\ref{fig:CrossSectionParameterization}. 

\begin{figure}[htp]
    \vspace{5px}
    \centering
    \includegraphics[width=7cm]{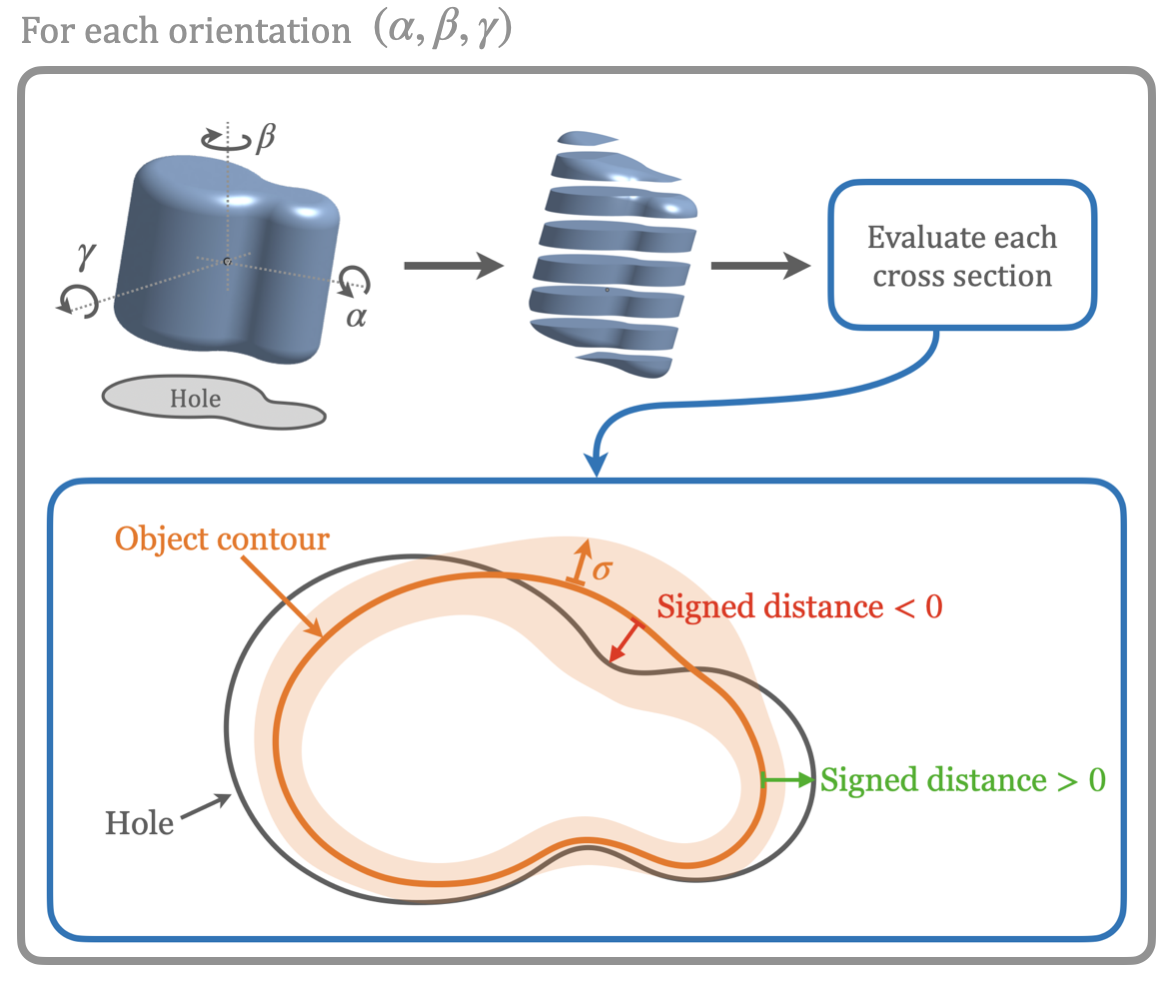}
    \caption{Overview of our object evaluation: For each orientation $(\alpha, \beta, \gamma)$, we split the object into a set of horizontal cross sections of uniform width. Each cross section is assigned a probabilistic score using Monte Carlo sampling, reflecting the expected interference between the hole and the cross section. This interference is measured as the minimum signed distance between the hole's contour and the sampled object contour.}
    \label{fig:ApproachOverview}
\end{figure}


Next, we evaluate the likelihood that each cross section will fit into the hole. We start by projecting the GP's cross section to the $x\textit{-}y$ plane. We parameterize the object's projected contour and the hole's contour with polar coordinates $\theta, r_{\text{obj contour}}(\theta)$ and $\theta, r_{\text{hole}}(\theta)$ respectively. In order for the object to fit into the hole, we need $r_{\text{obj contour}} < r_{\text{hole}}\ \forall\theta \in [-\pi, \pi]$. We define each section score as 
\begin{align}
    S_{\text{section}}(\alpha, \beta, \gamma, \l) = \displaystyle \min_{\theta} r_{\text{hole}}(\theta) - r_{\text{obj contour}}(\theta)
\end{align}

This score represents the worst location on the object for fit, and a negative score indicates that the object will not fit through the hole.
For a given object orientation, the likelihood that the object will fit into the hole is upper-bounded by the worst section, so we set the orientation score to be 
\begin{align}
    S_{\text{ori}}(\alpha, \beta, \gamma) = \displaystyle \min_l S_{\text{section}}(\alpha, \beta, \gamma, \l)
\end{align}

We use Monte Carlo sampling from the GP describing the object's surface, $R(\theta, \phi) \sim \mathcal{N}(\bar{f}, \mathbb{V}[f])$, to estimate the object's contour $r_{\text{obj contour}}(\theta) =  \displaystyle \max_\phi r(\theta, \phi) \cos(\phi)$ and obtain a probabilistic estimate of these scores.

We use these scores to select the next section of the object to explore. Given that each section is parameterized by $\alpha, \beta, \gamma, l$, we decompose this process into first selecting the best orientation $\alpha, \beta, \gamma$ and then selecting the height $l$.

Intuitively, we trade off between selecting an orientation that is most likely to position the object to fit through the hole according to the current model (exploitation), and the orientation with the most uncertainty about its fit (exploration). We can use the mean and standard deviation of $S$ to construct the following acquisition function, based on the {\em Upper Confidence Bound\/} (UCB) function~\cite{gpucb}: 
\begin{align}
\label{eq:ucb0}
    UCB(\alpha, \beta, \gamma) = \mu_{S_{\text{ori}}}(\alpha, \beta, \gamma) + \lambda \sigma_{S_{\text{ori}}}(\alpha, \beta, \gamma),
\end{align}
where $\lambda$ is a hyperparameter that controls the preference between exploitation and exploration. To select the next orientation, we maximize Equation~\ref{eq:ucb0} to obtain $\alpha^*, \beta^*, \gamma^*$. 

Next, we choose the horizontal section of the object in the selected orientation $(\alpha^*, \beta^*, \gamma^*)$. Here, we trade off between selecting the horizontal section that is most likely to overlap the hole (exploitation) and the horizontal section with the most uncertainty (exploration). This ensures that we tend to examine the `worst' sections of our `best' orientations. This is motivated by the need to examine the sections in the `best' orientation that are most likely to collide with the hole and cause task failure. If such `worst' sections still fit, the entire object in this orientation is likely to fit into the hole. Formally, we select the parameter $l^*$ (which defines the placement of the section to consider on the vertical axis) by maximizing the following function: 
\begin{align}
\label{eq:ucb1}
    UCB_{\alpha^*, \beta^*, \gamma^*}(l) = &-\mu_{S_{\text{section}}(\alpha^*, \beta^*, \gamma^*)}(l) \\
    &+ \lambda \sigma_{S_{\text{section}(\alpha^*, \beta^*, \gamma^*)}}(l) \notag
\end{align}

\subsection{In-hand reorientation}
We use the velocity controller from \cite{yuan2020design} to determine the rollers' pitch angles $\theta_{L,\text{pitch}}, \theta_{R,\text{pitch}}$ and rollers' angular velocities $\omega_{L,\text{roll}}, \omega_{R,\text{roll}}$ that are necessary to achieve the desired angular velocity of the object. The controller assumes that the object is a sphere, but we find that by adding compliance to the opening of the Roller Grasper, we can reorient a broader set of object shapes.

Due to torsional friction between the rollers and the object, when the roller grasper changes its pitch angle, the object may inadvertently rotate with the roller. This rotation is undesirable because it is hard to control the effects of torsional friction, and it prevents changing roller orientation relative to the object. To mitigate this, we use extrinsic dexterity \cite{dafle2014extrinsic} and lightly press the object against an external surface when changing the pitch angle of the rollers to prevent object rotation. 

\section{EXPERIMENTAL EVALUATION}
To demonstrate the effectiveness of our method, we present a quantitative evaluation in simulation. Additionally, we perform a noise analysis to guide future work in developing the next generation of Tactile-Enabled Roller Grasper. 

\subsection{Simulated task oriented exploration}

To evaluate our BO-based task guided exploration strategy, we measure the insertion task success rate and number of required exploration steps using 15 objects and 72 initial grasping points per object. An exploration step consists of a complete rotation of the object along a selected cross section. 

\textbf{Simulation environment.} The simulation environment is built in PyBullet \cite{pybullet}. First, an object is instantiated in-hand with a random orientation sampled from a uniform distribution. Next the rollers roll the object in-hand and collect tactile data. When a closed loop is detected, the roller stops and proceeds to the next section determined by the selected algorithm. The exploration process stops when the insertion error is less than 3mm. 

\begin{figure}[t]
    \vspace{7px}
    \centering
    \includegraphics[width=0.7\linewidth]{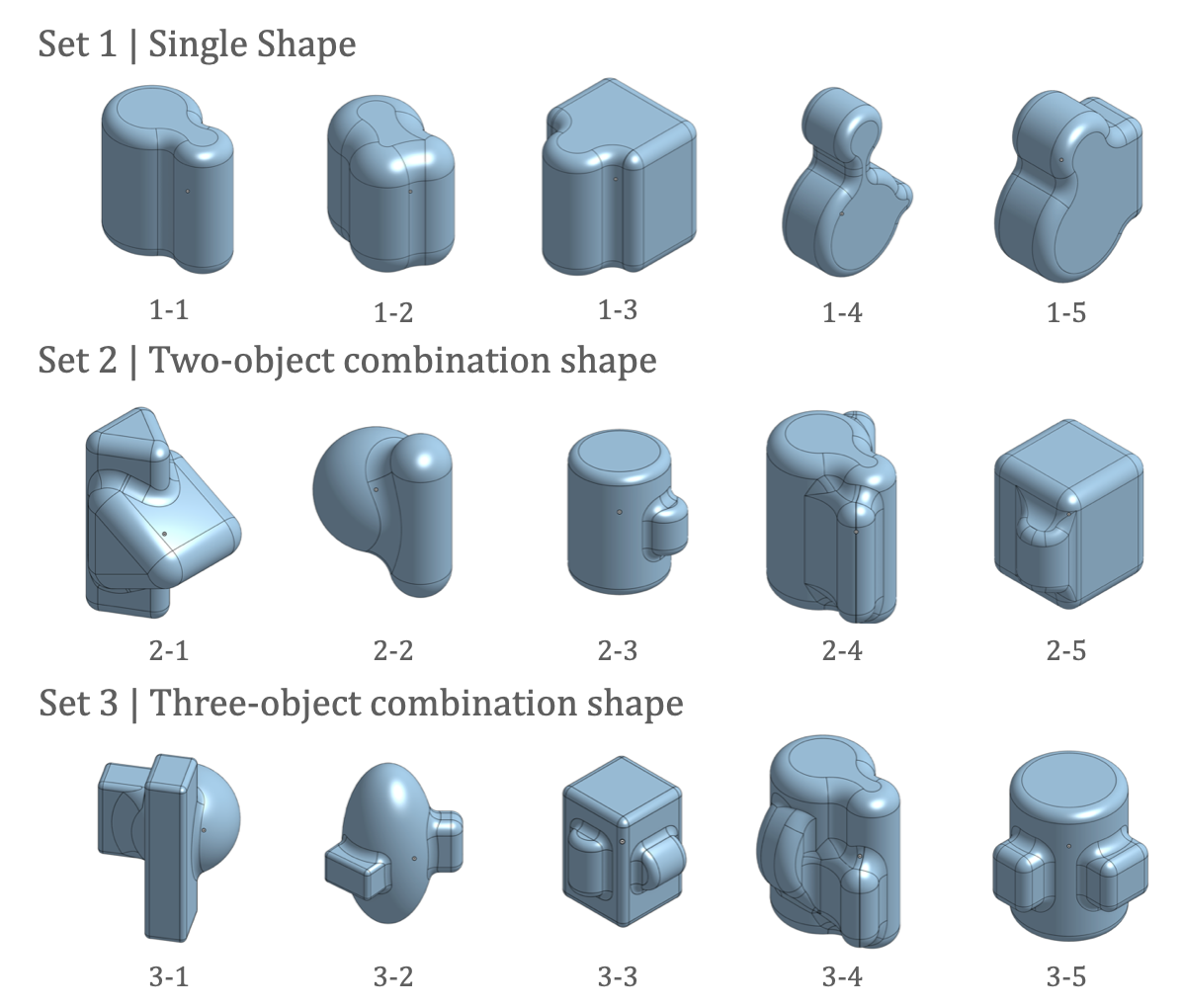}
    \caption{The object set used for evaluation. }
    \label{fig:exp:main:object}
    \vspace{-10px}
\end{figure}

\textbf{Object set.} Fig.~\ref{fig:exp:main:object} shows the 15 different objects we used for evaluation, generated by randomly combining either one, two, or three randomly-transformed basic 3D shapes. The target hole shape is generated by projecting the object onto a plane and verifying that the object can fit in the hole in only one orientation. This ensures that our algorithm must be accurately reconstructing the object shape to solve the task. The maximum object width is 5cm. 


\textbf{Methods we compare}.
\begin{enumerate}
    \item Our method that uses Bayesian optimization to trade off between exploration and task completion, with $\lambda$ optimized over a hold-out set of objects ($\lambda=500)$.
    \item Random:  selects sections for exploration from a uniform random distribution
    \item Exploit-only: without considering uncertainty, selects the object orientation that is most likely to allow insertion and picks the section in that orientation that is most likely to cause a failure. This is equivalent to our BO algorithm with $\lambda=0$. 
    \item Explore-only: selects the object orientation and section with the most uncertainty. This is equivalent to our BO algorithm with $\lambda \rightarrow \infty$. 
\end{enumerate}

The maximum exploration time allowed is 10 steps. 

\begin{figure}[]
    \centering
    \vspace{7px}
    \includegraphics[width=8cm]{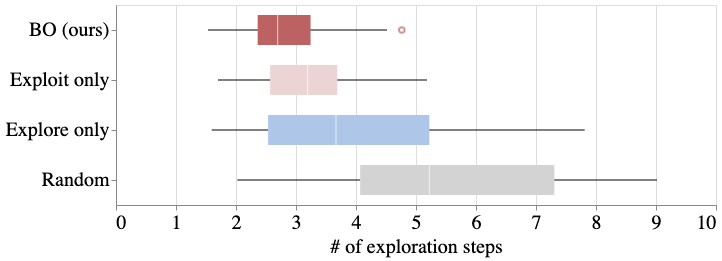}
    \caption{Number of exploration steps required for successful object insertion using different algorithms in simulation. The boxplot indicates the 25th percentile, median, and 75th percentile over the 15 objects tested.}
    \label{fig:exploration_step}
    \vspace{-15px}
\end{figure}

\textbf{Quantitative results.} Each result is evaluated over 15 objects with 72 initial grasping points. All methods are able to perform with a high success rate given enough exploration steps $>98\%$. However, our BO-based exploration strategy consistently requires fewer exploration steps (Fig.~\ref{fig:exploration_step}). Over the randomly-generated set of objects, our BO approach performs better than exploit-only ($p < 0.01$),  explore-only ($p < 0.01$), and the random baseline ($p < 0.001$). All p-values were computed using the paired-samples t-test.

\begin{figure}[t]
    \vspace{7px}
    \centering
    \includegraphics[width=0.9\linewidth]{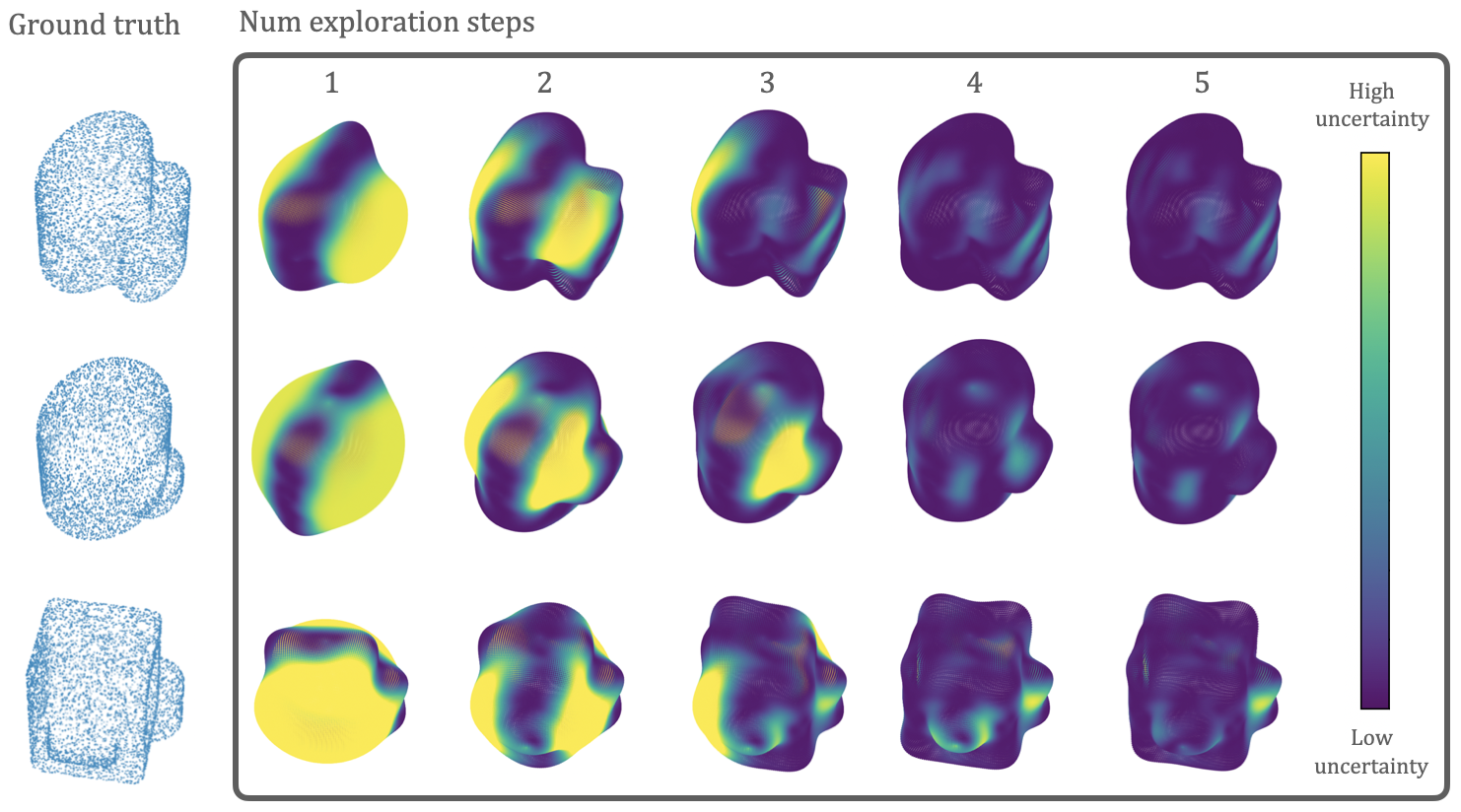}
    \caption{Qualitative visualization of the shape reconstruction.}
    \label{fig:shape_reconstruction}
    \vspace{-10px}
\end{figure}

\textbf{Shape reconstruction.} While shape reconstruction is \textit{not} the explicit objective of our algorithm, we still demonstrate that qualitatively, we end up recovering a meaningful approximation of global object shape as an auxiliary benefit of minimizing the task-driven objective (Fig.~\ref{fig:shape_reconstruction}).

\subsection{Sensitivity analysis for noise}

\begin{figure}[t]
    \centering
    \vspace{7px}
    \includegraphics[width=\linewidth]{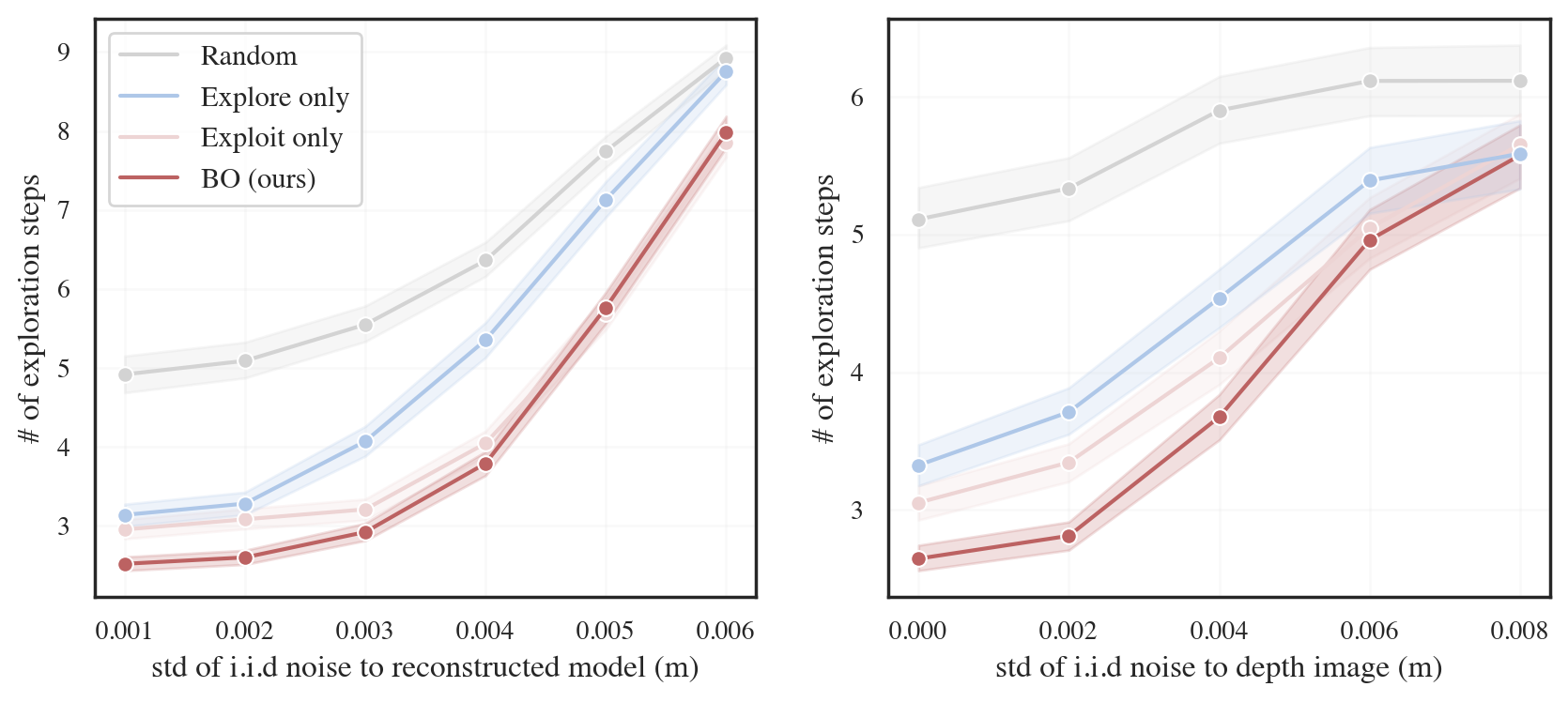}
    \caption{Steps to finish the exploration with i.i.d. noise. Each bar is evaluated over 5 objects with 72 different initial grasping points. }
    \label{fig:exp:sense:step_iid}
    \vspace{-10px}    
\end{figure}

\textbf{Noise in the reconstructed point cloud.} Noisy depth images and a noisy prior estimate on the object's motion due to slipping could cause poor object reconstruction results. To evaluate the tolerance to poor object reconstruction, we add i.i.d. zero-mean Gaussian noise to the reconstructed point cloud before fitting the GP model. As shown in Fig.~\ref{fig:exp:sense:step_iid} (left), while the success rate deteriorates in all methods as noise increases, our method outperforms others in the low-to-medium noise regimes. For high-noise regime our method performs similarly to the exploit-only baseline.

\textbf{Noise in the depth image from the tactile sensor} To evaluate the tolerance to noisy depth images, we add i.i.d. zero-mean Gaussian noise to the simulated depth images used for ICP  reconstruction. Unlike adding noise to the reconstructed model, adding noise to depth images makes it harder for ICP to reconstruct the object model. Fig.~\ref{fig:exp:sense:step_iid} (right) demonstrates that while adding noise to the depth images affects the performance of all methods, our method has a significant advantage in the low-to-medium noise regimes. When the standard deviation of the noise is smaller than 4 mm, our method can still reach a success rate above $80\%$. Higher noise levels can lead to a noticeable performance drop, at which point our method achieves comparable performance to our baselines. This is because our method depends on making informed guesses of where to explore next based on the model's current estimate of the object shape. With high levels of noise, the shape estimate becomes very inaccurate, and our algorithm no longer has enough information to make an informed guess of where to explore next. 

\textbf{Hardware implications} 
The simulation study shown here provides instructive feedback for deployment of our approach on the Tactile-Enabled Roller Grasper hardware. The Tactile-Enabled Roller Grasper used in this study is a novel gripper that is still under development. We empirically estimate the approximate noise of the depth image to be at least 7mm. Our simulation study (Fig.~\ref{fig:exp:sense:step_iid}) demonstrates that at this level of precision, the BO algorithm's benefit over a simpler strategy is limited. Our primary takeaway is that the gripper's sensor noise needs to be reduced by replacing the 3D printed plastic links with stiffer materials and reducing the backlash in the joints to improve the gripper's proprioception. Additionally, adding a third roller is needed to enable the rollers to change pitch angle without experiencing high shear forces that degrade the GelSight surface. The existing two roller design requires maintaining a high normal force when changing pitch angle to avoid dropping the object which causes high shear forces.

\section{CONCLUSION}
In summary, we present a method to reorient unknown objects with tactile sensing that does not rely on vision. We perform in-hand simultaneous 3D shape reconstruction and localization, and outline an efficient strategy based on Bayesian optimization to select regions of the object to explore to ensure quick task completion. We demonstrated the efficacy of this method on the insertion task, suggesting its possible broad utility in tactile manipulation. For example, our approach could be applied to assembly tasks of small objects, and reorientation of objects in cluttered spaces. While our method leverages the benefits of the Tactile-Enabled Roller Grasper, the method we propose on how to guide tactile exploration to be task-driven could apply to linkage-based grippers too. 

For success on the currently available hardware, our work relies on several assumptions. Because we use local tactile data, our approach requires objects to have heterogeneous surface features in order to localize the robot on the object and align neighboring tactile depth maps. The kinematic limitations of the Roller Grasper require that object shapes have aspect ratios close to 1. In future work, this limitation could be overcome by allowing regrasping or adding a third roller. Another complication is that objects may slip and drop in the early stages of exploration when the robot has very little object shape data or when the tactile measurements are too noisy. These limitations could be addressed via multimodal fusion with vision, which is a promising avenue for future work.






\bibliographystyle{unsrt}
\bibliography{citations}

\end{document}